\documentclass{article}
\usepackage{arxiv}
\usepackage{CJKutf8}
\usepackage[utf8]{inputenc} 
\usepackage[T1]{fontenc}    
\usepackage{hyperref}       
\usepackage{url}            
\usepackage{booktabs}       
\usepackage{amsfonts}       
\usepackage{nicefrac}       
\usepackage{microtype}      
\usepackage{lipsum}
\usepackage{graphicx}
\usepackage{amsmath}
\usepackage{float}
\usepackage{lscape}
\usepackage{fancyhdr}
\usepackage{graphicx}
\usepackage{subfigure}
\usepackage{indentfirst}
\usepackage{paralist}
\usepackage{enumitem}
\usepackage{threeparttable}

\usepackage[justification=centering]{caption}

\usepackage{paralist}

\graphicspath{ {./images/} }

\title{Yuan 1.0: Large-Scale Pre-trained Language Model in Zero-Shot and Few-Shot Learning}

\author{
 Shaohua Wu\thanks{wushaohua@inspur.com, Inspur Artificial Intelligence Research Institute}
   \And
 Xudong Zhao  
  \And
 Tong Yu \\ 
 \AND
 Rongguo Zhang 
  \And
 Chong Shen
  \And
 Hongli Liu 
  \And
 Feng Li \\
  \AND 
 Hong Zhu 
  \And
 Jiangang Luo
  \And
 Liang Xu
  \And
 Xuanwei Zhang
}

\begin{document}
\fancyhf{} 
\fancyhead[CO,CE]{Yuan 1.0: Large-Scale Pre-trained Language Model in Zero-Shot and Few-Shot Learning}

\maketitle
\begin{abstract}
Recent work like GPT-3 has demonstrated excellent performance of Zero-Shot and Few-Shot learning on many natural language processing (NLP) tasks by scaling up model size, dataset size and the amount of computation. However, training a model like GPT-3 requires huge amount of computational resources which makes it challengeable to researchers. In this work, we propose a method that incorporates large-scale distributed training performance into model architecture design. With this method, Yuan 1.0, the current largest singleton language model with 245B parameters, achieves excellent performance on thousands GPUs during training, and the state-of-the-art results on natural language processing tasks. A data processing method is designed to efficiently filter massive amount of raw data. The current largest high-quality Chinese corpus with 5TB high quality texts is built based on this method. In addition, a calibration and label expansion method is proposed to improve the Zero-Shot and Few-Shot performance, and steady improvement is observed on the accuracy of various tasks. Yuan 1.0 presents strong capacity of NLP tasks, and the generated articles are difficult to distinguish from the human-written ones.
\end{abstract}


\section{Introduction}
The Transformer architecture has been widely used in natural language processing\cite{ref1, ref3, ref4}. In order to improve the performance, a varieties of Transformer-based modifications have been proposed since 2017\cite{ref1}, but many of them exhibit a lack of generalization across different implementations and tasks\cite{ref27}. Kaplan, et al.\cite{ref2} confirms that performance of the Transformer steadily improves with the scaling up of model size, dataset size, and the amount of computation for training. Roberta\cite{ref3} shows that the accuracy of BERT can be substantially improved by training the model for a longer time with a larger corpus. The T5 model, built with vanilla Transformer structure and increased model size with 11 billion parameters, achieves the state-of-the-art (SOTA) performance in various NLP tasks\cite{ref4}. It is proved that larger language models performs better than smaller ones.\\
GPT-3 with 175 billion parameters, as a milestone, was proposed in 2020\cite{ref5}. Before GPT-3, it was common to pre-train a model with unsupervised learning on a large unlabeled dataset, then fine-tune on a specific task. Because GPT-3 makes great progress on Zero-Shot and Few-Shot learning, it can be applied directly on a wide range of NLP tasks, and displays good performance without  being fine-tuned on those tasks. After GPT-3, several studies further increase the model size in two ways:
\begin{enumerate}[$\bullet$]
    \item Singleton: Increase the number of layers and the size of a layer, such as GPT-3 and PanGu-$\alpha$\cite{ref6}.
	\item Mixture of Experts (MoE): Scaling the model size with Sparsely Gated Mixture-of-Experts (MoE), such as GShard\cite{ref7}, Switch Transformer\cite{ref8}, Wudao\cite{ref9, ref10} and M6\cite{ref11} . Each expert is a singleton model in a size up to 10B. With MoE, the model size can be successfully scaled up to more than 1000B \cite{ref8,ref10}.
\end{enumerate}
Both Singleton and MoE are effective to increase the model size, however, they behave differently in Zero-Shot and Few-Shot scenarios. Currently, the MoE method still follows the common way, in which pre-train the model on a large dataset and fine-tune it on specific task. To our best knowledge, no MoE model is applied on Zero-Shot or Few-Shot learning. However, both GPT-3 and PanGu-$\alpha$ with singleton architecture, exhibits good performance on Zero-Shot and Few-Shot learning\cite{ref5,ref6}.\\
Training a model with parameters greater than 100B requires huge amount of computational resources. Take GPT-3 175B for example, it was trained on a cluster of 10,000 GPUs \cite{ref5,ref28}. Such a huge requirement on computational resources makes it difficult for most researchers to train a model in a similar way. In this work, we propose Yuan 1.0 singleton model with 245B parameters. To accelerate the training process of Yuan 1.0, and thus reduce energy costs and carbon emissions, we make a collaborative design of model architecture and large-scale distributed training. The main contributions of our work are summarized as below,
\begin{enumerate}[$\bullet$]
    \item A method that incorporates large-scale distributed training performance into model architecture design is proposed. With this method, we trained our Yuan 1.0, the current largest singleton language model with 245B parameters, and achieved excellent performance on thousands GPUs
	\item A data processing system is created to efficiently filter a massive amount of data from Internet. The current largest Chinese corpus with 5TB high-quality text is built based on this system.
	\item The model architecture with better performance in Pre-train and Fine-tune pattern is likely to behave opposite in Zero-Shot and Few-Shot learning.
	\item A method that can steadily improve the Zero-Shot and Few-Shot performance is proposed.  
\end{enumerate}
\section{Yuan 1.0}
The basic architecture of Yuan 1.0 is a language model. For a given input $(x_1, x_2, \ldots, x_n)$, the language model predicts the probability of the output $(y_1,y_2,\ldots, y_n)$:
\begin{align}
    p(y) = \prod_{i=1}^n p(y_i | x_1, x_2, \ldots, x_{i-1}) \label{eq1}
\end{align}
To deal with different downstream tasks (translation, question answering, classification etc.), each task is casted in a diagram of text-to-text framework\cite{ref4}. In this way, a pre-trained language model can be directly applied to handle different tasks\cite{ref4,ref12}.
\subsection{Model architecture}
\begin{figure}[ht]
    \centering
    \subfigure[LM]{
    \begin{minipage}[t]{0.4\linewidth}
    \centering
    \includegraphics[width=6cm]{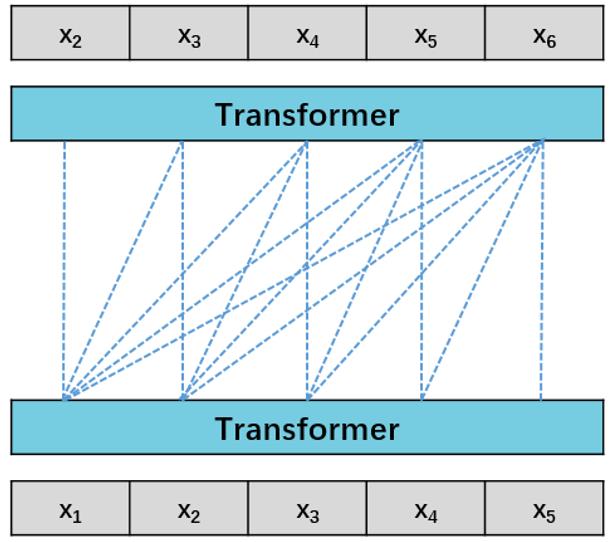}
    \label{Figure 1 (a)}
    \end{minipage}
    }
    \subfigure[PLM]{
    \begin{minipage}[t]{0.4\linewidth}
    \centering
    \includegraphics[width=6cm]{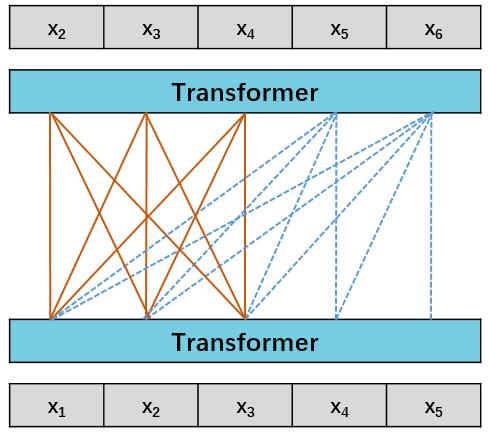}
    \label{Figure 1 (b)}
    \end{minipage}
    }
    \caption{Schematics of (a) LM and (b) PLM. Solid orange lines stand for fully-visible mask in prefix portion of input, and blue dot lines stand for casual mask.}
    \vspace{-0.4cm}
    \label{Figure 1}
\end{figure}

In this work, we consider two model architectures, Language Model (LM) and Prefix Language Model (PLM). In LM, which is one of the most commonly used architecture\cite{ref5,ref12,ref13}, the decoder in a Transformer is taken to auto-regressively generate an output sequence. The structure of a decoder LM is presented in Fig. \ref{Figure 1 (a)}. \\
At time t, a token at the rightmost of the output, $x_6$, is generated based on the probability predicted by the model. Then this token is concatenated to the input sequence and fed back into the model to generate the next token at time t+1. In the pre-train and fine-tune pattern, a LM performs better in natural language generation (NLG) tasks, but comparatively worse in natural language understanding (NLU) tasks. The reason of this drawback is that a LM, with casual masking of attention structure, forces the model’s prediction of the $i^{th}$ token only depends on the tokens before i, which can be seen in Fig. \ref{Figure 1 (a)}. In contrast, PLM performs well in both NLU tasks and NLG tasks\cite{ref4}. Instead of taking a casual mask, PLM uses a fully-visible mask in the range of the prefix portion of input sequence, which can be seen in Fig. \ref{Figure 1 (b)}.
\subsection{Cooperative design of model structure}
The huge cost of computational resources is the bottlenecks that limits researchers to develop NLP models with hundreds of billions parameters. Take GPT-3 for example, it was trained on a large cluster with 10,000 GPUs\cite{ref5,ref28}. In order to accelerate the training process, we incorporate the key factors that affect the performance of large-scale distributed training into the design of Yuan 1.0 model structure.  \\
The parameters of LM that affects both the accuracy and the performance of large-scale distributed training include the number of layers, hidden size, global batch size, micro batch size, etc. In the large-scale distributed training of Yuan models, we use three-dimensional parallel strategies, including tensor parallelism, pipeline parallelism and data parallelism\cite{ref14}. In this section, we make theoretically analysis of these parameters, and demonstrate how to choose the model parameters under different parallel strategies. The notations used are presented in Table \ref{Table 1}.
\begin{table}[H]
    \centering
    \begin{tabular}{l|l}
     \toprule
     Symbol & Notation\\
     \midrule
     {p} & {The size of pipeline parallelism.}\\
     {t} & {The size of tensor parallelism.}\\
     {d} & {The size of data parallelism.}\\
     {n} & {The number of GPUs. $p \cdot t \cdot d = n$}\\
     {B} & {Global batchsize}\\
     {b} & {Micro-batchsize}\\
     {m} & {The number of micro batches in a model pipeline group,$m=\frac{1}{b}\cdot \frac{B}{d}$}\\
     {L} & {The number of Transformer layers}\\
     {h} & {Hidden size}\\
     {S} & {Sequence length}\\
     {l} & {The number of layers in each pipeline stage, $l=L/p$}\\
     \bottomrule
    \end{tabular}
    \setlength{\abovecaptionskip}{0.2cm}
    \caption{Notations in theoretical analysis of model parameters.}
    \vspace{-0.6cm}
    \label{Table 1}
\end{table}
\subsubsection{Tensor Parallelism}
\begin{figure}[ht]
  \centering
  \includegraphics[width=14cm]{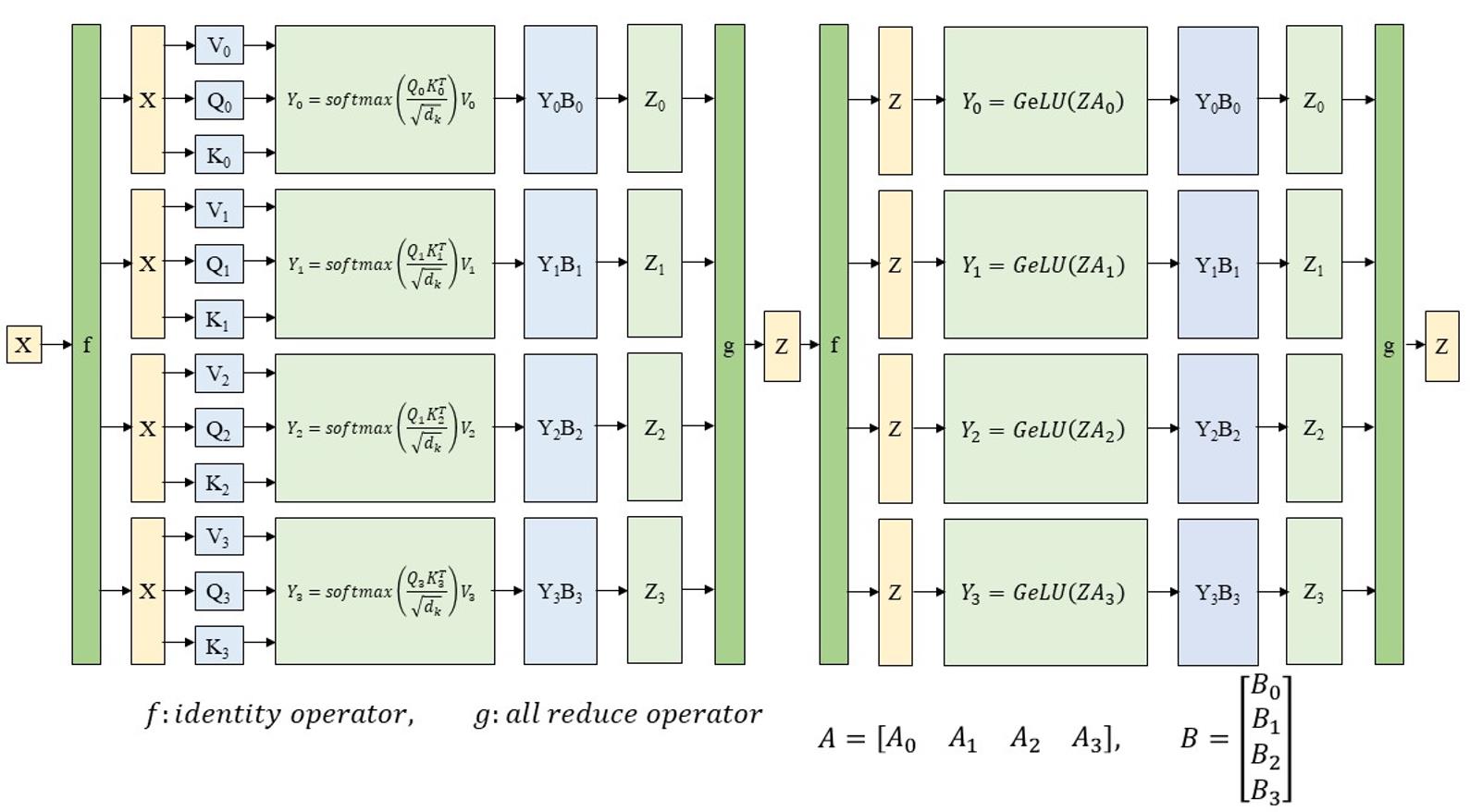}
  \caption{Schematics of tensor parallelism.}
  \vspace{-0.6cm}
  \label{Figure 2}
\end{figure}
In tensor parallelism, the layers of a model are partitioned among devices within a node. The schematic of tensor parallelism is presented in Figure \ref{Figure 2}. In Transformer, tensors of Attention and MultiLayer perceptron (MLP) are split by row or column during forward and backward computing. Input tensor is broadcasted to each accelerator, in which makes forward propagation. When the forward pass of Attention or MLP is finished, an all-reduce is performed. Then the results are updated on all devices and sent to the next Layer. There are four all-reduce operations in the forward and backward propagation per layer.  \\
The ratio of the computation time to data communication time per layer $f_{tp}$ is,
\begin{align}
    f_{tp} = \frac{96t}{8 \left( t-1 \right)} \left( h + \frac{S}{6} \right) \label{eq2}
\end{align}
According to Eq. \ref{eq2}, $f_{tp}$ of the tensor parallelism increases with h and S. The value of S is usually chosen as 512, or 1024 \cite{ref4,ref6}. Because memory requirements of Attention is quadric to S, sparse Attention structure is necessary if S is increased to 2048\cite{ref5}. In order to save memory of accelerators and enable larger S, the activations are recomputed in backward propagation\cite{ref15}. With this method, the model can be trained with $S=2048$ smoothly with normal Attention structure.

\subsubsection{Pipeline Parallelism}
\begin{figure}[ht]
  \centering
  \includegraphics[width=14cm]{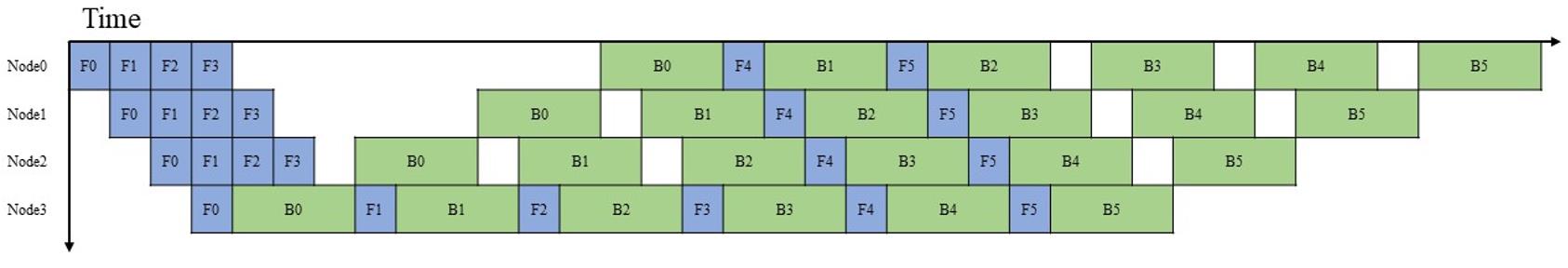}
  \caption{Schematics of pipeline parallelism. F0, $\ldots$, F3 means forward pass in pipeline stage 0, $\ldots$, stage 3, while B0, $\ldots$, B3 means backward pass in pipeline stage 0, $\ldots$, stage 3.}
  \vspace{-0.2cm}
  \label{Figure 3}
\end{figure}
For language models with hundreds of billions parameters, the parameters can hardly be stored in a single node. Pipeline parallelism spliting the layers of LM among multiple nodes, is applied to solve the above mentioned problem (Fig. \ref{Figure 3}). Each node is one stage in the pipeline, which receives outputs from the previous stage and sends results to the next one. A node will be idle if the inputs received from its previous neighbor is not ready. The idle time for a pipeline is called pipeline bubble\cite{ref14,ref16}. To increase the performance of pipeline parallelism, we have to decrease the time spent on pipeline bubble. The fraction of ideal time spent in the pipeline bubble $f_{pb}$ is,
\begin{align}
    f_{pb}=\frac{(L/l-1)}{m} \label{eq3}
\end{align}
According to Eq. \ref{eq3}, the time spent on pipeline bubble increases with the number of layers L, and decreases with the number of micro-batch size $m$. There will be a better performance if $m \gg L/l$. In pipeline parallelism, the ratio of the computation time to data communication time per node $f_{pp}$ is, 
\begin{align}
    f_{pp}=24\frac{L}{p} \left( h+\frac{S}{6} \right) \label{eq4}
\end{align}
According to Eq. \ref{eq4}, the computational efficiency of a pipeline node improves with the increase of the values of h and S, which is similar to the situation of tensor parallelism.
\vspace{-0.1cm}
\subsubsection{Data Parallelism}
\begin{figure}[ht]
  \centering
  \includegraphics[width=14cm]{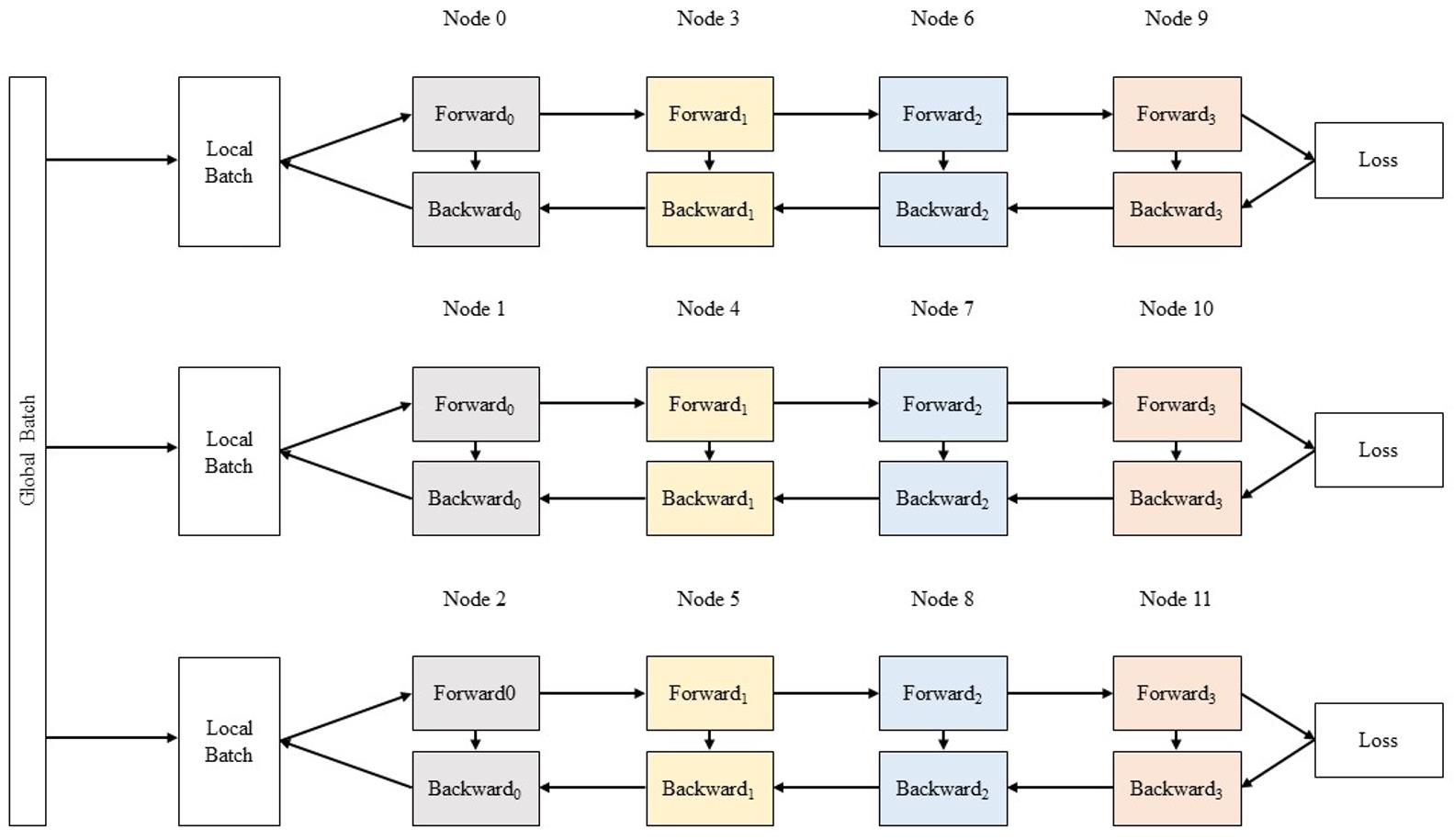}
  \caption{Schematics of data parallelism.}
  \vspace{-0.6cm}
  \label{Figure 4}
\end{figure}
The global batch size is split among pipeline groups by data parallelism (Fig. \ref{Figure 4}).  Each pipeline group with a copy of the model is fed by local batches. In data parallelism, the ratio of computing time to communication time $f_{dp}$ is,
\begin{align}
    f_{dp}\approx \frac{4BSd}{d-1} \label{eq5}
\end{align}
Because d is often far greater than 1, Eq. \ref{eq5} can be simplified to,
\begin{align}
    f_{dp} \approx 4BS \label{eq6}
\end{align}
The computing efficiency improves with the increase of the global batch size B and sequence length S. Because the memory requirements is quadric to sequence length S, increasing the global batch size seems to be a more effective way. However, there will be numerical instabilities during training when global batch size is too large \cite{ref2,ref17}. To avoid numerical divergence, the global batch size is kept smaller than $10^7$ tokens.
\subsubsection{The principles of model parameters selections}
In summary, we follow the rules below to select model parameters,
\begin{enumerate}[$\bullet$]
    \item Increase the sequence length as much as possible, as it benefits the tensor parallelism, pipeline parallelism, and data parallelism. Because the memory requirement is quadric to the sequence length, it is worthy to re-compute activations in the backward propagation to save memories. 
    \item Too many layers in language model have negative effect in performance, because it increases the time spent on pipeline bubble. 
    \item Increasing the hidden size improves the performance of both tensor parallelism and pipeline parallelism.
    \item Increasing the number of micro batches in a node improves the performance of pipeline parallelism. Increasing the global batch size improves the performance of data parallelism.
\end{enumerate}
Three models (Yuan LM-13B, Yuan 13B-PLM and Yuan 245B) are trained with parameters presented in Table \ref{Table 2}.
\begin{table}[H]
    \centering
    \begin{tabular}{p{2.4cm}|p{1cm}|p{1.1cm}|p{1.1cm}|p{0.9cm}|p{1.5cm}|l|l|l|l|p{1.8cm}|p{1.4cm}}
     \toprule
     Model & Layers & {Hidden\newline size} & Global BS & Micro BS & Sequence Length & t & p & d & GPUs & Parameters (billion) & Training tokens (billion) \\
     \midrule
     {Yuan LM-13B} & {40} & {5120} & {2688} & {4} & {2048} & {8} & {2} & {112} & {1792} & {12.87} & {300} \\
     {Yuan PLM-13B} & {40} & {5120} & {2688} & {4} & {2048} & {8} & {2} & {112} & {1792} & {12.87} & {300} \\
     {Yuan 245B} & {76} & {16384} & {3360} & {1} & {2048} & {8} & {38} & {7} & {2128} & {245.73} & {180} \\
     \bottomrule
    \end{tabular}
    \setlength{\abovecaptionskip}{0.2cm}
    \caption{Parameters of Yuan models.}
    \vspace{-0.8cm}
    \label{Table 2}
\end{table}
\section{Dataset}
A Chinese corpus with 5TB high-quality text is built, which is sufficient to train Yuan 245B model without sampling the dataset twice. To our best knowledge, this is the largest Chinese text dataset compared with CLUECorpus2020 (100GB) \cite{ref18}, PanGu Corpus (1.1TB) \cite{ref6}, WuDaoCorpus2.0 (2.3TB Chinese text data and 300GB English text data) \cite{ref10}, and ERNIE 3.0 (4TB) \cite{ref19}.
\begin{figure}[ht]
  \centering
  \includegraphics[width=14cm]{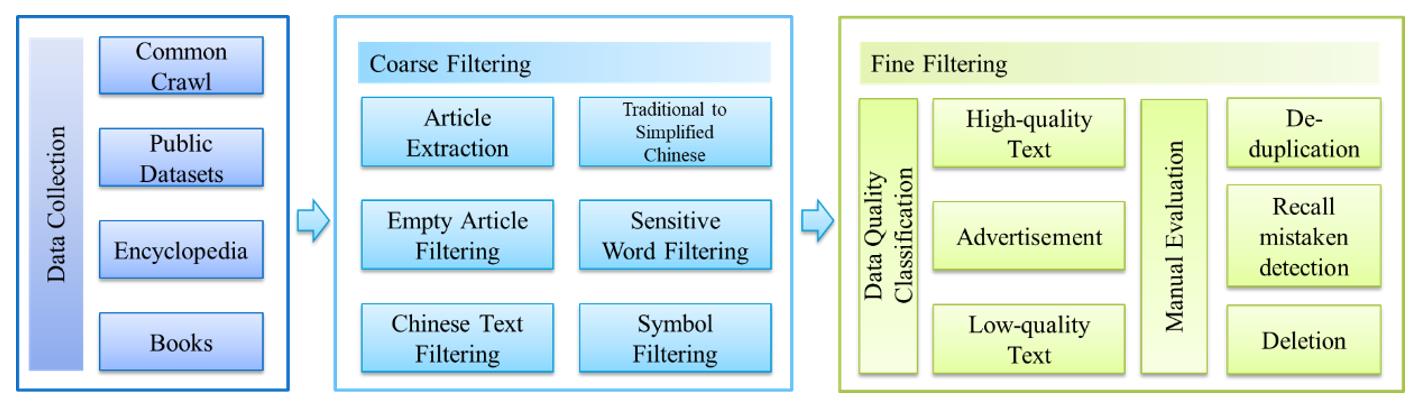}
  \caption{Procedure of dataset processing.}
  \vspace{-0.4cm}
  \label{Figure 5}
\end{figure}
In order to obtain the high-quality dataset, we develop a Massive Data Filtering System (MDFS) built on Spark to clean and filter the raw data, and train a Bert-based model to select high quality samples. MDFS is consisted of three parts, data collection, coarse filtering and fine filtering (Fig. \ref{Figure 5}). The raw data is collected from Common Crawl, Sogou News, SogouT, Encyclopedia, and Books (Table \ref{Table 3}). To process these raw data, we run MDFS system on a high performance cluster with 36 nodes.
\begin{table}[ht]
    \centering
    \begin{tabular}{l|l|l}
     \toprule
     Category & Size (GB) & Data source\\
     \midrule
     Common Crawl & 866,304 & Web pages from 2017 to 2021 \\
     Public datasets & 16,797 & SogouT, Sogou News \\
     Encyclopedia & 37 & Baidu Baike, Wikipedia \\
     Books & 4,020 & Books in philosophy, history, humanities, novel, etc.\\
     \bottomrule
    \end{tabular}
    \setlength{\abovecaptionskip}{0.2cm}
    \caption{Composition of raw data.}
    \vspace{-1.2cm}
    \label{Table 3}
\end{table}
\subsection{Coarse Filtering}
The coarse filtering includes the following modules:
\begin{enumerate}[$\bullet$]
    \item Article Extraction: Extracts contents of articles from crawled web pages, and removes field like WARC headers, hyperlinks, etc. The following rules are applied,
    \begin{enumerate}[a)]
        \item A paragraph started with WARC keyword is discarded.
        \item A paragraph without a valid punctuation at the end is discarded.
        \item A paragraph that contains neither English character nor Chinese character is discarded.
    \end{enumerate}
    \item Empty Article Filtering: Removes empty articles.
    \item Chinese Text Filtering: Selects Chinese texts with the following rules,
    \begin{enumerate}[a)]
        \item An article with less than 30 Chinese characters is removed.
        \item An article is removed if the percentage of Chinese characters is less than 60.
        \item After this step, the data size of Common Crawl is decreased from 866,304GB to 12,200GB.    
    \end{enumerate}
    \item Traditional to Simplified Chinese: Converts traditional Chinese characters to simplified Chinese characters.
    \item Sensitive Words Filtering: Removes articles or paragraphs that include sensitive words. 9,759 sensitive words are collected and classified into black and blue categories.
    \begin{enumerate}[a)]
        \item An article with words in black category is removed.
        \item A paragraph that contains words in blue category is removed, and other paragraphs in the same article is kept.
    \end{enumerate}
    \item Symbol filtering: Removes junk symbols, such as invisible Unicode characters, invisible ASCII characters and specific punctuations.
\end{enumerate}
\vspace{-0.4cm}
\subsection{Fine Filtering}
To extract high quality articles based on course filtering text, we train a Bert-based model to classify high quality, low quality and advertisements. A datasets labeled with high quality articles, low quality articles, and advertisements is built to train this model.\\
About 2TB data is removed by the model, and 50\% of the removed data are identified as advertisements. Considering that the advertisements may also contain complete semantic information, we evaluate them manually to determine whether it is necessary to recall. The processed dataset scatters on 36 nodes. To avoid additional bias from human, we sample two sets on each node, and each set is evaluated by different reviewers. Parts of the statistical results are shown in Table \ref{Table 4}.
\begin{table}[ht]
    \centering
    \begin{tabular}{l|l|l}
     \toprule
     server ID & Sample1 & Sample2\\
     \midrule
     {cc01} & {1.25} & {0.60} \\
     {cc02} & {2.17} & {1.85} \\
     {cc03} & {2.88} & {3.43} \\
     {cc04} & {6.59} & {6.48} \\
     {cc05} & {1.02} & {1.49} \\
     {cc06} & {1.08} & {1.13} \\
     {cc07} & {0.64} & {0.72} \\
     {cc08} & {0.70} & {0.86} \\
     {cc09} & {0.65} & {0.60} \\
     {cc10} & {0.94} & {1.04} \\
     {cc11} & {0.70} & {0.81} \\
     {cc12} & {0.40} & {0.39} \\
     {mean} & {1.59} & {1.62} \\
     \bottomrule
    \end{tabular}
    \setlength{\abovecaptionskip}{0.2cm}
    \caption{Percentage of high-quality data in advertisements on different servers.}
    \vspace{-0.2cm}
    \label{Table 4}
\end{table}
The similar percentage of high quality data for Sample1 and Sample2 indicates a high consistency in data evaluation. As the percentage of high-quality data in advertisements is fairly low, it is reasonable to discard all advertisements. In the manual review, we find a 2.4\% duplication rate in high-quality contents, while the duplication rate is 12.6\% in advertisements. De-duplication is further applied to the high quality data. \\
The data size after fine filtering is shown in Table 5. The total size of high-quality dataset is 5.02TB. During training, only articles with more than 150 characters are sampled.
\begin{table}[H]
    \centering
    \begin{tabular}{l|l|l}
     \toprule
     Category & Size before (GB) & Size after (GB)\\
     \midrule
     {Common Crawl} & {866,304} & {4,200} \\
     {Public datasets} & {16,797} & {268} \\
     {Encyclopedia} & {37} & {10.5} \\
     {Books} & {4,020} & {655} \\
     \bottomrule
    \end{tabular}
    \setlength{\abovecaptionskip}{0.2cm}
    \caption{Size of high-quality data after fine filtering.}
    \vspace{-0.4cm}
    \label{Table 5}
\end{table}
\section{Experiments and Results}
The Yuan models are trained on a cluster with 2128 GPUs. A stable real performance of 45\% of the theoretical peak performance is achieved on this cluster. Adam optimizer is used for training Yuan models. Please refer to Table 6 for more details. To stabilize the training process, a linear warm up of learning rate is taken over the first 1\% tokens, then the learning rate follows a cosine curve that slowly decays to 10\% of its original value. The global batch size also linearly increases to the full value over the first 2\% tokens, then it is kept till the end of training. During the training we pack multiple documents into a single sequence of size 2048, and the documents are separated with a special token "<eod>". The tokens in the sequence are not masked in any way. 
\begin{table}[H]
    \centering
    \begin{tabular}{l|l|l|l|l}
     \toprule
     Models & Learning Rate & Weight Decay & $\beta_1$ in Adam & $\beta_2$ in Adam\\
     \midrule
     {Yuan LM-13B} & {1.0E-4} & {0.01} & {0.9} & {0.95} \\
     {Yuan PLM-13B} & {1.0E-4} & {0.01} & {0.9} & {0.95} \\
     {Yuan 245B} & {1.6E-4} & {0.1} & {0.9} & {0.95} \\
     \bottomrule
    \end{tabular}
    \setlength{\abovecaptionskip}{0.2cm}
    \caption{Hyper-parameters for Yuan models.}
    \vspace{-0.6cm}
    \label{Table 6}
\end{table}
\subsection{Tasks description}
The models are mainly evaluated on FewCLUE and ZeroCLUE\cite{ref18}, which can be classified into 4 categories, including text classification, Winograd Schema, natural language inference, and reading comprehension.\\
\textbf{Text Classification} is consisted of sentimental classification (Eprstmt: E-commerce Product Review Dataset for Sentiment Analysis), news title classification (Tnews: Toutiao Short Text Classification for News), app description classification (Iflytek: Long Text classification), and subject classification (Csldcp: Chinese scientific literature subjects classification).\\
Eprstmt is a binary classification with positive and negative product reviews. Tnews, Iflytek and Csldcp are multi-class classification with 15, 118 and 67 categories respectively. On tasks with labels as 0 or 1 or in English, we assign each label with a semantically Chinese meaningful name. For labels longer than one token, we convert those into one-token labels with the same meaning. For all text classification tasks, label is appended to the end of a sentence, connected with prompt words. Our generative model predicts the label based on a given sentence, and calculate the probability P(label|sentence)  of each candidate. The candidate with the largest probability is selected as the prediction.\\
\textbf{Winograd Schema task (Wsc)} is a disambiguation task determining which noun a pronoun refers to. It is treated as a binary classification task in our evaluation.\\
\textbf{Natural Language Inference (NLI)} includes Ocnli and Bustm, which concerns the ability to understand the relation between two sentences. Both of the tasks provide two sentences and a label of 0 or 1. We calculate the cross entropy loss of the second sentence with a candidate label, and treat the label with the lowest loss as the prediction.\\
\textbf{Reading Comprehension} includes Chid\cite{ref20} and Csl. Chid is a Chinese Idiom cloze test dataset. Each sentence has a blank inside, and for each blank, there are 7 candidate idioms with 1 true choice. For this task, each candidate is filled in the blank, and we calculate the cross entropy loss of each combination. The one with the lowest loss is the predicted true idiom. Csl can be treated as either a reading comprehension or a binary classification task. An abstract is provided along with 4 keywords in the dataset. If all keywords are consistent with the abstract, the label should be true or 1. Otherwise, the label should be false or 0. All keywords are appended to the end of abstract. We calculate the cross entropy loss of the part after ABSTRACT in condition of ABSTRACT. The one with smaller loss is the predictive result.\\
\textbf{Gereration tasks}: In addition to FewCLUE and ZeroCLUE, there are two generation tasks, CMRC2018 and WebQA. CMRC2018 is a span text extraction task, consisted of articles followed by several questions, and the answer to a question is a segment of the corresponding article. WebQA is a closed book question answering task. Model is evaluated by directly answering questions of CMRC2018 and WebQA without any auxiliary information. The generated answer is evaluated with EM and F1 scores.
\vspace{-0.4cm}
\subsection{Comparison of LM and Prefix LM}
Yuan LM-13B and Yuan PLM-13B are evaluated on FewCLUE\cite{ref21} and ZeroCLUE\cite{ref18} (Table \ref{Table 7}). The SOTA results of ZeroCLUE is benchmarked with zero-shot, while that of FewCLUE are benchmarked with fine-tune. The Zero-Shot results are in-context learning without tuning parameters.
\begin{table}[ht]
    \vspace{-0.2cm}
    \centering
    \begin{tabular}{l|l|l|l|l|l|l|l|l|l|l}
     \toprule
     & {Scores} & {Bustm} & {Chid} & {Csl} & {Csldcp} & {Eprstmt} & {Iflytek} & {Ocnli} & {Tnews} & {Wsc}\\
     \midrule
     {Human} & {82.48} & {88} & {87.1} & {84} & {68} & {90} & {66} & {90.3} & {71} & {98} \\
     {SOTA} & {49.881} & {69.25} & {60.85} & {50.63} & 26.41 & 78.09 & 25.69 & 37.04 & 52.93 & 51.03 \\
     {LM-13B} & {56.88} & {59.375} & {86.14} & {50} & {47.533} & {88.125} & {37.87} & {46.875} & {57.01} & {38.99} \\
     {Prefix LM-13B} & {55.83} & {56.875} & {85.63} & {48.13} & {46.57} & {88.125} & {38.82} & {48.125} & {57.468} & {32.7} \\
     \bottomrule
    \end{tabular}
    \ \ \ \\
    \ \ \\
    \ \ \\
    \begin{tabular}{l|l|l|l|l|l|l|l|l|l|l}
     \toprule
     & {Scores} & {Bustm} & {Chid} & {Csl} & {Csldcp} & {Eprstmt} & {Iflytek} & {Ocnli} & {Tnews} & {Wsc}\\
     \midrule
	 {Human} & {82.48} & {88} & {87.1} & {84} & {68} & {90} & {66} & {90.3} & {71} & {98} \\
     {SOTA} & {70.16} & {76.7} & {69.45} & {76.57} & {59.55} & {88.05} & {45.77} & {72.02} & {73.87} & {73.1} \\
     {LM-13B} & {69.14} & {81.25} & {72.27} & {81.25} & {60.9} & {90.625} & {54.87} & {46.25} & {52.45} & {82.39} \\
     {Prefix LM-13B} & {72.66} & {83.75} & {76.24} & {83.75} & {66.44} & {90.625} & {57.78} & {58.125} & {56.74} & {80.5} \\
     \bottomrule
    \end{tabular}
    \setlength{\abovecaptionskip}{0.2cm}
    \caption{Performance of Yuan models on ZeroCLUE and FewCLUE tasks. The results are measured on evaluation dataset. (a)Performance of Yuan models on ZeroCLUE tasks with Zero-Shot learning (top); (b) Performance of Yuan models on FewCLUE with Fine-tune (bottom). }
    \vspace{-0.8cm}
    \label{Table 7}
\end{table}
Table \ref{Table 7}(a) indicates both LM and PLM have convincing in-context learning capability. The zero-shot average scores of both LM and PLM are superior to the SOTA one. On Csldcp, Tnews and Iflytek tasks, we surpass the zero-shot SOTA by a large margin. Our models also achieve strong performance on Ocnli, which is 6-8 points larger than the zero-shot SOTA. Table \ref{Table 7}(a) displays the results after calibration and label expansion, and the methods in details will be discussed in the next section. \\
Our supervised fine-tuning method is aligned with the design of GPT\cite{ref12}. The average scores for LM and PLM are comparable to the SOTA ones (Table \ref{Table 7}(b)). Compared to the few-shot learning results, fine-tune makes great improvement to Bustm, Csl and Wsc. However, for Chid, Eprstmt, Tnews and Ocnli, which are strong in zero-shot, fine-tune contributes little or even have negative effect. The fine-tuned accuracy is superior to the SOTA fine-tune results on 7 tasks including Bustm, Chid, Csl, Csldcp, Eprstmt, Iflytek and Wsc.
\begin{table}
    \centering
    \begin{tabular}{l|l|l|l|l|l|l|l|l|l|l}
     \toprule
     & {Scores} & {Bustm} & {Chid} & {Csl} & {Csldcp} & {Eprstmt} & {Iflytek} & {Ocnli} & {Tnews} & {Wsc}\\
     \midrule
	 {Human} & {82.48} & {88} & {87.1} & {84} & {68} & {90} & {66} & {90.3} & {71} & {98} \\
	 {SOTA} & {49.881} & {69.25} & {60.85} & {50.63} & {26.41} & {78.09} & {25.69} & {37.04} & {52.93} & {51.03} \\
	 {LM-13B} & {59.024} & {57.5} & {87.5} & {51.66} & {47.92} & {84.99} & {34.58} & {45.22} & {64.47} & {62.76} \\
     \bottomrule
    \end{tabular}
    \ \ \ \\
    \ \ \\
    \ \ \\
    \begin{tabular}{l|p{1cm}|l|l|l|l|l|l|l|l|l}
     \toprule
     & {Scores} & {Bustm} & {Chid} & {Csl} & {Csldcp} & {Eprstmt} & {Iflytek} & {Ocnli} & {Tnews} & {Wsc}\\
     \midrule
	 {Human} & {82.48} & {88} & {87.1} & {84} & {68} & {90} & {66} & {90.3} & {71} & {98} \\
	 {SOTA} & {70.16} & {76.7} & {69.45} & {76.57} & {59.55} & {88.05} & {45.77} & {72.02} & {73.87} & {73.1} \\
	 {PLM-13B} & {71.451} & {72.05} & {87.4} & {80.53} & {62.75} & {88.45} & {47.31} & {45.12} & {68.33} & {87.93} \\
     \bottomrule
    \end{tabular}
    \setlength{\abovecaptionskip}{0.2cm}
    \caption{Performance of Yuan models on (a) ZeroCLUE (top) and (b) FewCLUE (bottom) tasks. The results were evaluated by the CLUE server.}
    \vspace{-0.4cm}
    \label{Table 8}
\end{table}
We submitted the PLM on FewCLUE, and LM on ZeroCLUE. Both of them currently topped on the list (Table \ref{Table 8}). Comparing the results of LM and PLM, we note that LM performs better on Zero-Shot learning, while PLM outperforms with fine-tune. Fine-tune in general brings better accuracy in most tasks. However, fine-tune costs tremendous computational resources for Yuan 245B model, which makes fine-tune uneconomic. Accordingly, we choose LM as basic architecture of Yuan 245B model. 
\subsection{Results of Yuan 245B model}
\begin{figure}[ht]
  \centering
  \includegraphics[width=14cm]{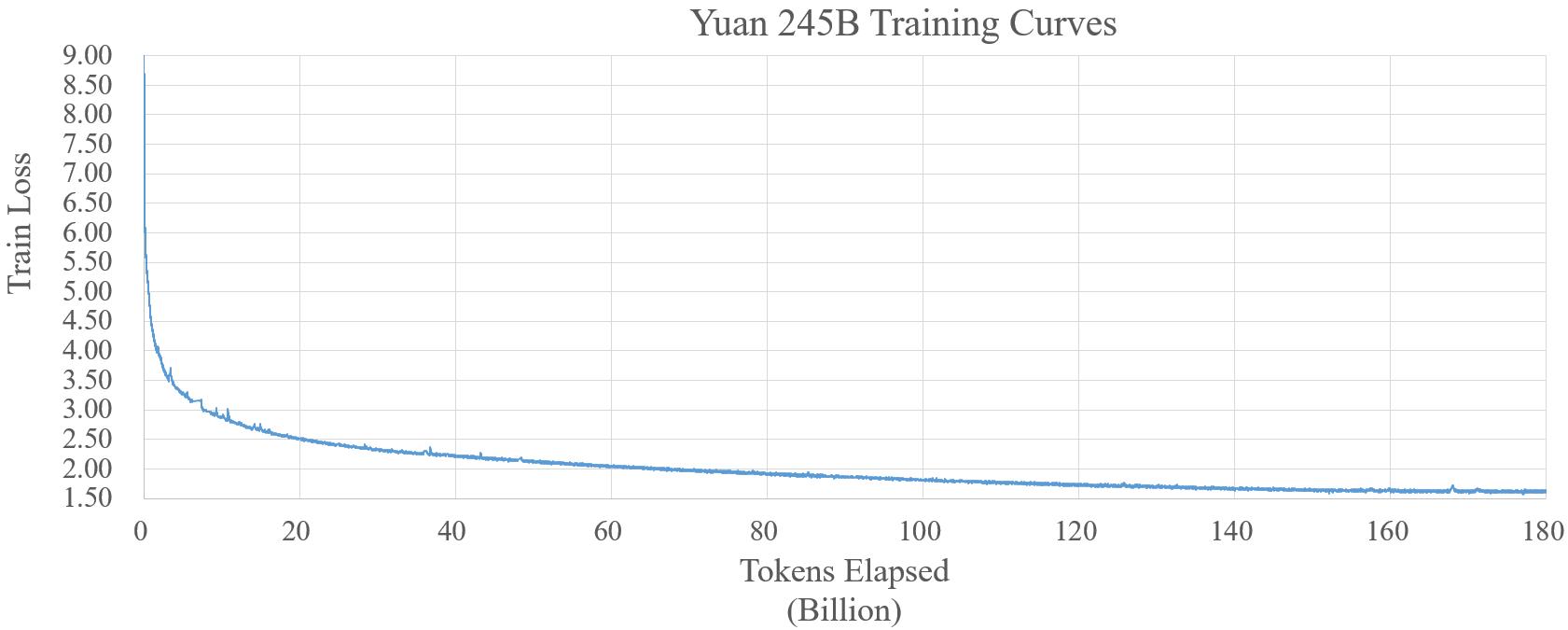}
  \caption{Training loss of Yuan 245B model.}
  \label{Figure 6}
\end{figure}
\begin{table}[H]
    \centering
    \begin{tabular}{l|l|l|l}
     \toprule
     Models & Number of Parameters & PetaFlops-days & Training loss\\
     \midrule
     {GPT-3} & {175B} & {3640} & {1.73} \\
     {PanGu-$\alpha$} & {200B} & {796.3} & {2.49} \\
     {Yuan 245B} & {245B} & {4095} & {1.64} \\
     \bottomrule
    \end{tabular}
    \setlength{\abovecaptionskip}{0.2cm}
    \caption{Comparison of GPT-3, PanGu-$\alpha$ and Yuan.}
    \vspace{-0.8cm}
    \label{Table 9}
\end{table}
Fig. \ref{Figure 6} presents the training loss curves of Yuan 245B model. The loss decreases rapidly at the first 10B tokens, and gets flatten over a long tail. \\
Table \ref{Table 9} shows the comparison of GPT-3, PanGu-$\alpha$ and Yuan 245B in training. The PetaFlops-day of PanGu-$\alpha$ and Yuan is computed as, 
\begin{align}
    \frac{8 \times Number\_of\_tokens \times Number\_of\_parameters}{8.64E19} \label{eq7}
\end{align}
Activations are recomputed during backward propagation. The computing amount of Yuan 245B is much greater than that of PanGu-$\alpha$. The training loss of Yuan 245B is the smallest among these three models. \\
Table \ref{Table 10} compares the generation results between Yuan and recently published Chinese pretrained language models, Pangu-$\alpha$\cite{ref6} and Ernie 3.0\cite{ref19}. The average scores of Yuan outperform Pangu-$\alpha$ and Ernie 3.0 by a large margin on Close-book QA\cite{ref22} and Span Extraction reading comprehension\cite{ref23}, which proves the excellent zero-shot generation capacity of Yuan 245B. Regarding WebQA, Yuan significantly improves the performance, no matter evaluated with EM or F1 Score. For CMRC2018, Yuan also achieved a better averaged score and F1 score compared to the SOTA, and it is little worse on EM compared to the SOTA.
\begin{table}[H]
    \centering
    \begin{minipage}{1.2\textwidth}
    \begin{minipage}[t]{0.45\textwidth}
    \begin{tabular}{l|l|l|l}
     \toprule
     WebQA & Score & EM & F1\\
     \midrule
     PanGu-$\alpha$ & 9.8 & 5.13 & 14.7 \\
     Ernie 3.0 (SOTA) & 30.74 & 22.53 & 38.95 \\
     Yuan 245B & 40.467 & 30.57 & 50.36 \\
     \bottomrule
    \end{tabular}
    \end{minipage}
    \begin{minipage}[t]{0.45\textwidth}
    \begin{tabular}{l|l|l|l}
     \toprule
     CMRC2018 & Score & EM & F1\\
     \midrule
     PanGu-$\alpha$ & 10.37 & 1.46 & 19.28 \\
     Ernie 3.0 (SOTA) & 16.61 & 7.61 & 25.61 \\
     Yuan 245B & 27.37 & 5.58 & 49.17 \\
     \bottomrule
    \end{tabular}
    \end{minipage}
    \end{minipage}
    \setlength{\abovecaptionskip}{0.2cm}
    \caption{Performance of Pangu-$\alpha$, Ernie 3.0, and Yuan 245B model on WebQA (left) and CMRC2018 (right) tasks.}
    \vspace{-0.6cm}
    \label{Table 10}
\end{table}
\subsubsection{Accuracy improvements on X-Shot learning}
A noticeable shortcoming of in-context learning lies in its bias towards template sentences and labels. The bias mainly comes from a dataset with imbalance distribution between classes, few-shot examples with a certain order, and labels with different frequencies in the training corpus\cite{ref24}. The extraneous bias limits model’s performance on natural language tasks. Considering the probable sources of bias, we take calibration for in-context learning in two aspects: a calibration on the calculation of probability, and an expansion  of labels.\\
Based on previous work on calibration\cite{ref24,ref25}, the model’s bias can be fixed with an empty text. We use a similar method to calibrate our in-text prediction. 
\begin{align}
    \arg \max \frac{P(prediction | given\ sentence)}{P(prediction | void)} \label{eq8}
\end{align}
Take Tnews and Ocnli as examples. In the case of Tnews, we calculate the probability of the last token in the sentence-label combination, which is actually a prediction of label.\\\
\begin{CJK*}{UTF8}{gbsn}
Orig：新闻：sentence。这条新闻是关于label。\\
Void：新闻：N/A。这条新闻是关于label。\\
With calibration, we calculate,
\begin{align}
    \arg \max \frac{P(label | \mbox{新闻：sentence。这条新闻是关于})}{P(label | \mbox{新闻：N/A。这条新闻是关于})} \label{eq9}
\end{align}
In the case of OCNLI, which is a two-sentence task, we calculate the cross entropy loss of the second sentence.\\
Orig: sentence1? 对/错/可能，sentence2.\\
Void：N/A？对/错/可能，sentence2.\\
With calibration, we calculate:
\begin{align}
    \arg \max \frac{P(label | \mbox{sentence1?对/错/可能})}{P(label | \mbox{N/A?对/错/可能})} \label{eq10}
\end{align}
\end{CJK*}
In the experiment, we find that the difference in label frequency are influential to the prediction. The ideal condition is that all labels have approximately the same frequency, but it is too tough for manual selection. We choose an Embedding Corpus\cite{ref26} that covers over 8 million Chinese words and phrases, to assist us reveal the relevance between words. Each label is expanded into 5 synonyms, in order to reduce the bias led by a single label word or phrase.\\
Three tasks are selected to display the effect of calibration, and the results are presented in Table \ref{Table 11}. The combination of model calibration and label expansion  leads a dramatic improvement for Eprstmt, which is a binary sentimental analysis. For multi classification on news (Tnews) and scientific literature (Csldcp), calibration also brings better results to a large extent.
\begin{table}[H]
    \centering
    \begin{tabular}{l|l|l|l}
     \toprule
     & {Csldcp} & {Eprstmt} & {Tnews} \\
     \midrule
     {Original} & {37.96} & {54.3} & {53.28} \\
     {Calibration and label expansion} & {48.017} & {86.88} & {57.19} \\
     \bottomrule
    \end{tabular}
    \setlength{\abovecaptionskip}{0.2cm}
    \caption{Accuracy improvement of Yuan 245B model by calibration and label expansion. The results were evaluated on ZeroCLUE evaluation datasets.}
    \vspace{-0.6cm}
    \label{Table 11}
\end{table}
\begin{figure}[ht]
  \centering
  \includegraphics[width=14cm]{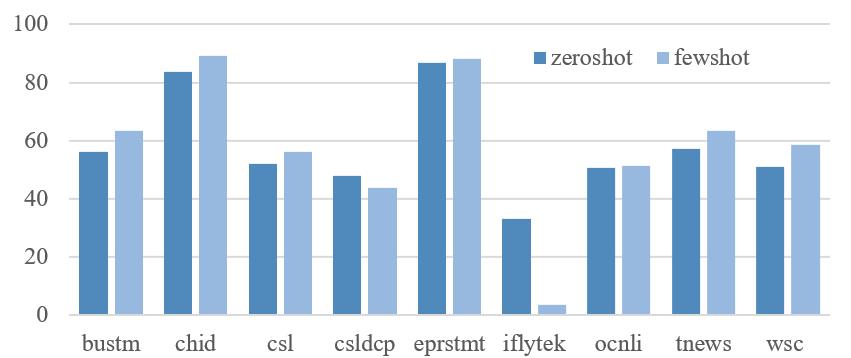}
  \caption{Performance of Yuan 245B in Zero-Shot and Few-shot learning.}
  \vspace{-0.4cm}
  \label{Figure 7}
\end{figure}
Figure \ref{Figure 7} presents the Zero-Shot and Few-Shot results of Yuan 245B on ZeroCLUE tasks. In Few-Shot, the number of samples is determined by the number of classes. We take 4 samples for binary classification, and 3 samples otherwise. Compared with zero-shot, we note that few-shot learning brings steady improvements on accuracy for most tasks, except Csldcp and Iflytek. Few-shot leads to a disastrous decrease for Iflytek with a near-random result. With the same method, few-shot has positive effect for Eprstmt (2 classes) and Tnews (15 classes). We observe similar situations in Yuan LM-13B, and Yuan PLM-13B. Because there are large number of classes in Csldcp (67 classes) and Iflytek (118 classes), and the samples cannot cover all the classes, the bias caused by samples concatenated in the input make the model predication getting worse.
\subsubsection{Text generation of Yuan 245B model}
In order to see how well Yuan can generated text, we arbitrarily select 24 articles that Yuan 1.0 generates, including 4 couplets, 5 traditional and modern Chinese poetries, 5 news articles, 5 stories and 5 dialogues. Couplet, poetry and dialogue can be seen as short-text task (~10-20 tokens), while news and story generation can be seen as long-text task (~300 tokens). In comparison, the human-written articles are masterpieces from Chinese poems, pieces of classic novels, news articles from Sohu News, and dialogues from LCCC-large dataset\cite{ref29}. Participants are asked to select whether the article was "written by a human" or "written by a model". We collect 83 valid questionnaire. According to our interview, most interviewee will choose "the better one" as the article created by human.\\
\begin{figure}[H]
  \centering
  \vspace{-0.6cm}
  \includegraphics[width=14cm]{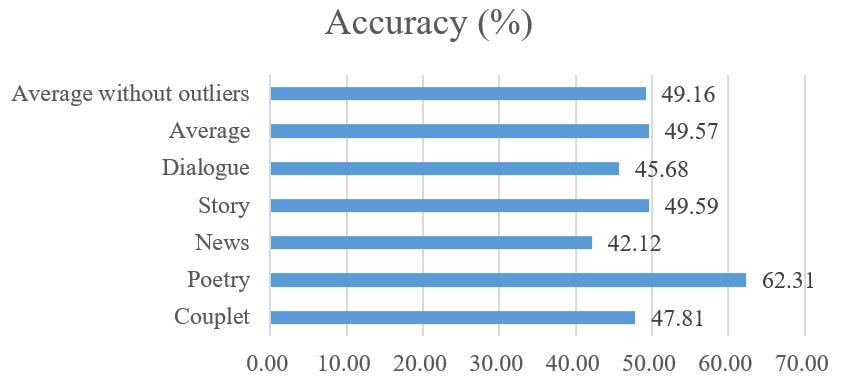}
  \caption{Human accuracy at detecting articles written by Yuan 245B model.}
  \label{Figure 8}
\end{figure}
\begin{landscape}
\begin{CJK*}{UTF8}{gbsn}
\begin{table}
    \centering
    \begin{tabular}{p{1.5cm}|p{2cm}|p{7cm}|p{7cm}|l}
     \toprule
      & Input & Human & Model & Accuracy \\
     \midrule
     Dia-logue & 我好心问你吃了饭没，你又骂我 & 宝贝儿我错了，再也不敢了 & 我又没骂你，你自己对号入座 & 38.28\% \\
     Poetry & 作一首七言诗 & {塞上长城万里长，\newline 漫天风雪映重峦。\newline 一边江水人心去，\newline 满目青山绝照还。} & {满目青山夕照明，\newline 愿随江舟泛轻风。\newline 今生常忆多情事，\newline 香闺如梦领神会。} & 65.44\% \\
     Couplet & 上联：春来人入画，下联： & 福至梦开花 & 夜半月当灯 & 38.28\% \\
     Story & {起床穿衣，洗漱完毕，我走下楼去。心情甚至有些愉快地向科芬道了早安。} &      {起床穿衣，洗漱完毕，我走下楼去。心情甚至有些愉快地向科芬道了早安。\newline 尽管这家伙昨天跟我开了个不大不小的玩笑。\newline 酒吧里已经聚了很多人了，住店的客人都来了。昨天没来得及细看，今天一注意才知道，这些人绝大多数都是捕鲸者：大副、二副、三副；铁匠、木匠、铜匠；还有标枪手、守船人，不一而足。\newline 他们的肤色一律棕黑，衣着随便，蓄胡子的人占了很大的比例。\newline 你可以通过他们的外表来判断他们在岸上已经呆了多长时间了。\newline 瞧，这个小伙子，两颊赤红，像烤过的梨，他从印度洋回来不超过三天。\newline 他旁边那一位，颜色没他那么重，身上似乎有点锻木的味道，他上岸有一个星期了。\newline 有的人脸上只剩下一丝隐隐约约的热带的黄色了，他们在岸上肯定已经呆了好几个星期了。\newline 不过，谁也不能从魁魁格的面色来判断他上岸的时间。}  & {却没想到，下一刻，我就受到了一个不应该出现在这里的人的冷待。 \newline 那人一身笔挺的深色西装，靠在车身上，一双灰蓝色的眼眸，充满了不满与敌意，打量着我。 \newline 科芬几乎是立刻挡在了我面前，警戒地看着眼前的人，道： “特里。 ” \newline 我眨了眨眼，大脑一时有些空白。 \newline 我努力地思考，到底发生了什么事。 \newline 这个人，应该是叫特里没错吧。 \newline 他应该是莫里的哥哥，但是他为什么会出现在这里？他不是应该在中央星吗？ \newline 他为什么要用这种眼神看着我？ \newline 我小心地打量着特里，不明白他的敌意是从哪里来的。 \newline  “科芬，你先去上班吧。 ”我说道。 \newline  “我知道了，林恩先生。 ”科芬看了特里一眼，才一步三回头地走向悬浮车。 \newline 我在心底叹息了一声，走上前去，伸出手，道： “你好，特里，我是林恩。 ” \newline 特里面无表情地看了我一眼，伸出手与我交握。} & 40.75\% \\
     {News} & {摘要：辽宁主汛期首日遭暴雨侵袭 当地多举措防洪涝灾害。正文：中新网沈阳7月22日电(司晓帅)} & {辽宁省22日进入主汛期,一场自西向东的区域性暴雨过程也随之而来。记者当日下午从辽宁省防汛抗旱指挥部获悉,该省已采取多项举措应对此次降雨过程,严防洪涝灾害的发生。据辽宁省气象部门介绍,从7月21日8时至22日14时,该省西部降中到大雨,局部大暴雨,其它地区降小到中雨。西部地区平均降雨量达65毫米,最大点雨量为葫芦岛地区要路沟站达150毫米,100毫米以上点雨量站达17处,50毫米以上点雨量站160处。} & {	记者22日从辽宁省防汛抗旱指挥部办公室获悉，今年主汛期首日，辽宁遭到强降雨袭击，全省平均降雨量达39毫米，为历年同期最高。其中，沈阳、大连、鞍山、抚顺、本溪、丹东、锦州等8个市平均降雨量在40毫米以上，最大降雨量出现在本溪市，达70.5毫米。据辽宁省气象部门预测， 22日夜间到23日，辽宁全省还将有一次大到暴雨过程。} & 34.15\% \\
     \bottomrule
    \end{tabular}
    \caption{Samples generated by human and Yuan 245B model.}
    \vspace{-0.8cm}
    \label{Table 12}
\end{table}
\end{CJK*}
\end{landscape}
The human accuracy at detecting articles created by Yuan 1.0 is 49.16\%, which implies the difficulty for participants to distinguish human-written and model-written articles, especially in regard to modern Chinese poetries and articles (Fig. \ref{Figure 8}). The generation of news (42.12,\%) and stories (49.15\%) convinces us with excellent long-text generation capacity. Some of model-written articles are even better than parts of masterpieces in the view of our participants. The generation of couplets and poetries indicate that Yuan 1.0 is able to create text with rules and forms of ancient Chinese (Table \ref{Table 12}) although ancient Chinese is not strengthened in our pretraining corpus. Yuan can also make a dialogue aligned with human’s expectation (45.68\%). Yuan is able to generate articles, such as news, and stories,  which is hard to tell whether the article is human-written or model-written. However, you will see repetition to some extent, if the model is required to create an article with more than 1000 Chinese characters.\\
Few-shot is also effective for text-generation, especially text with a certain format. Table \ref{Table 13} shows the couplets Yuan created with zero-shot, one-shot and three-shot, under the same set of hyper-parameters. Few-shot mainly contributes to: (a) increasing the stability and completeness of generation (No. 3); (b) more reasonable semantic meaning and accordant written style (No. 1); (c) avoidance of the word repetition that is a taboo for couplet creation (No. 2, 5); and (d) antithetically better (No. 4).
\begin{CJK*}{UTF8}{gbsn}
\begin{table}
    \centering
    \begin{tabular}{p{2cm}|p{3cm}|p{3cm}|p{3cm}|p{3cm}}
     \toprule
      No & Input & Zero-Shot & One-Shot & Three-Shot \\
     \midrule
     1 & 好事流传千古 & 红眼盯住一人 & 高风亮节万年 & 佳音永驻万家 \\
     2 & 清风明月自来往 & 白鹤白鹿随去来 & 好鸟好花留我住 & 白鹤黄莺长往还 \\
     3 & 三江顾客盈门至 & 四海宾朋 & 四海财源进宝来 & 四海宾朋满座来 \\
     4 & 和风吹绿柳 & 小鸟唱好诗 & 细雨点红桃 & 细雨点红梅 \\
     5 & 三阳日照平安宅 & 三阳居财丁兴旺 & 五福星临吉庆门 & 五福星临福寿门 \\
     \bottomrule
    \end{tabular}
    \setlength{\abovecaptionskip}{0.2cm}
    \caption{Couplets generated by Yuan 245B with Zero-Shot, One-Shot and Three-Shot.}
    \vspace{-0.4cm}
    \label{Table 13}
\end{table}
\begin{table}
    \centering
    \begin{tabular}{p{7cm}|p{7cm}}
     \toprule
     input & Output \\
     \midrule
     “悬命”是拼命，接近全力，做出最大努力的意思。他凡事必悬命，绝不打马虎眼。用“悬命”造句： & 他这个人做事太悬命了，为了取得第一手材料，连饭也顾不上吃 \\
     “腕前”与本领的意思相同。例：“他只使出了了一小部分腕前，就把事情完成得如此出色”。用“腕前”造句： & 这个人的腕前很大，他很能完成这项任务 \\
     气持”是心情、精神状态的意思。例：“那种消极的气持又控制了他”。用“气持”写一句话： & 他的气持，使我无法接近 \\
     “油断”意思是漫不经心，马虎，轻率。举例：“她虽然在戏中只扮演一个次要的角色，但她并没有半点油断，精彩的表演得到大家的称赞”。请用“油断”写一个句子： & 他的学习成绩很好，因为他并没有半点油断 \\
     “胜手”意思是最得意拿手的，擅长的。比如，“别的饭食我不会做，只有做面条还胜手”。请用“胜手”一词造句： & 我的作文水平在班里名列前茅，是同学们公认的胜手  \\
     \bottomrule
    \end{tabular}
    \setlength{\abovecaptionskip}{0.2cm}
    \caption{The ability of imitation of Yuan.}
    \vspace{-0.8cm}
    \label{Table 14}
\end{table}
\end{CJK*}
The ability of imitation of Yuan is evaluated via learning and utilizing a brand new word. A definition and an example sentence are given in the input, and the model will write a new sentence with the given information. The nonexistent words include nouns and adjectives. Table \ref{Table 14} displays the one-shot examples. In all cases, the model makes approximately correct applications with the nonexistent Chinese words, which implies the learning and imitation ability of our model. This ability is especially effective when model aids scientific article writing, as tremendous definition in academic articles could be alien to Yuan.\\
In few-shot and text generation experiments, we note the pre-trained language model’s sensitivity to steering, which could be a potential risk on its application. Given an input without bias, the opinion created by the model could be either positive or negative. However, if given an input with a strong standpoint, the model tends to continue the article with the same opinion. In the case displayed in Table \ref{Table 15}, we would like to talk about the women status in the society, and two contrasting options are given as the input. Given "traditional patriarchy still holds sway", Yuan express the opinion as "women status is determined by their fertility". On the opposite, given "40\% of the labor force are women. There are more readable girls and women than ever", the model also follows the opinion as "women have the competence to any work". It is proved that a model with no certain bias can be easily steered by human. As the capability of generating "human-written-style" article strengthen the risk in model abuse, the application of model needs to be regulated.
\begin{CJK*}{UTF8}{gbsn}
\begin{table}[H]
    \centering
    \begin{tabular}{p{5cm}|p{5cm}|p{5cm}}
     \toprule
     & Negative & Positive \\
     \midrule
     Input & 由于种种历史原因，社会中父权制文化根深蒂固。父权制文化将现实存在的性别歧视看作是自然形成的、理所当然的和不可改变的，是由妇女的生理状况决定的。这种观念已经渗透到了社会文化之中。社会公众和一些政治家的言论中充斥着对妇女的陈规定型的偏见。20世纪50年代至60年代，“让妇女返回家庭”成为一股强大的社会思潮。人们普遍认为，妇女的角色就是生儿育女，成为贤妻良母。关于男性和女性在社会地位上是否存在的差异，我认为 & 过去四分之一世纪里女孩和妇女的生活发生了巨大的变化。现在识字的女孩和妇女比历史上任何时候都多，而且在三分之一的发展中国家，学校里的女生人数多于男生。妇女现在占全球劳动力的 40\%。而且，在全世界所有地区，女性的平均寿命都超过了男性。这些变化发生的速度是惊人的，关于男性和女性在社会地位上是否存在的差异，我认为 \\
     Output & 现在已经不需要再做任何的争论了。 我相信，女性的社会地位和她们的生育能力之间有着直接的联系。女性的生育能力是社会地位的基础。女性在家庭中的地位是由她们的生育能力决定的。如果女性不能生育，那么她们在家庭中的地位也将下降。 & 这个问题应该由社会来决定，而不是由妇女来决定。 我认为，女性在工作中的表现与男性一样，她们有能力胜任任何工作。在现实生活中，她们中的很多人工作得比男性更出色。我希望我的女儿将来能够成为一名出色的律师、总统候选人、医生、总统国家安全事务顾问，或者宇航员。但是，我希望她不要因为别人告诉她不可以做什么而轻易放弃自己的理想。\\
     \bottomrule
    \end{tabular}
    \setlength{\abovecaptionskip}{0.2cm}
    \caption{The generation of Yuan 245B by positive and negative inputs.}
    \vspace{-0.8cm}
    \label{Table 15}
\end{table}
\end{CJK*}
\section{Conclusion}
We proposed the current largest singleton language model Yuan 1.0 with 245B parameters that achieved good performance on different NLP tasks in Zero-Shot and Few-Shot learning. The architecture of Yuan 1.0 was designed by incorporating model structure with key factors that affects performance of large-scale distributed training. The training process achieved excellent performance on 2128 GPUs. Yuan 1.0 was trained on a new Chinese dataset of 5TB high-quality text that was built on 850TB raw data from Internet. Zero-Shot and Few-Shot performance was steadily improved by calibration and label expansion. We found that the prefix language model that performs better in Pre-train and Fine-tune pattern behaves differently in Zero-Shot learning, while language model behaved the opposite. Yuan 1.0 models achieved the state-of-the-art results on ZeroCLUE, FewCLUE, and generation tasks. The articles generated by Yuan 1.0 are difficult to distinguish from those written by humans.
\bibliographystyle{ieeetr}  
\bibliography{Yuan.bib}

\begin{thebibliography}{10}

\bibitem{ref1}
A.~Vaswani, N.~Shazeer, N.~Parmar, J.~Uszkoreit, L.~Jones, A.~N. Gomez,
  {\L}.~Kaiser, and I.~Polosukhin, ``{Attention is all you need},'' {\em
  Advances in Neural Information Processing Systems}, vol.~2017-Decem,
  pp.~5999--6009, jun 2017.

\bibitem{ref3}
Y.~Liu, M.~Ott, N.~Goyal, J.~Du, M.~Joshi, D.~Chen, O.~Levy, M.~Lewis,
  L.~Zettlemoyer, and V.~Stoyanov, ``{RoBERTa: A Robustly Optimized BERT
  Pretraining Approach},'' 2019.

\bibitem{ref4}
C.~Raffel, N.~Shazeer, A.~Roberts, K.~Lee, S.~Narang, M.~Matena, Y.~Zhou,
  W.~Li, and P.~J. Liu, ``{Exploring the limits of transfer learning with a
  unified text-to-text transformer},'' {\em Journal of Machine Learning
  Research}, vol.~21, 2020.

\bibitem{ref27}
S.~Narang, H.~W. Chung, T.~Yi, W.~Fedus, and C.~Raffel, ``Do transformer
  modifications transfer across implementations and applications?,'' 2021.

\bibitem{ref2}
J.~Kaplan, S.~McCandlish, T.~Henighan, T.~B. Brown, B.~Chess, R.~Child,
  S.~Gray, A.~Radford, J.~Wu, and D.~Amodei, ``{Scaling Laws for Neural
  Language Models},'' {\em arXiv}, 2020.

\bibitem{ref5}
T.~B. Brown, B.~Mann, N.~Ryder, M.~Subbiah, J.~Kaplan, P.~Dhariwal,
  A.~Neelakantan, P.~Shyam, G.~Sastry, A.~Askell, S.~Agarwal, A.~Herbert-Voss,
  G.~Krueger, T.~Henighan, R.~Child, A.~Ramesh, D.~M. Ziegler, J.~Wu,
  C.~Winter, C.~Hesse, M.~Chen, E.~Sigler, M.~Litwin, S.~Gray, B.~Chess,
  J.~Clark, C.~Berner, S.~McCandlish, A.~Radford, I.~Sutskever, and D.~Amodei,
  ``{Language Models are Few-Shot Learners},'' 2020.

\bibitem{ref6}
W.~Zeng, X.~Ren, T.~Su, H.~Wang, Y.~Liao, Z.~Wang, X.~Jiang, Z.~Yang, K.~Wang,
  X.~Zhang, C.~Li, Z.~Gong, Y.~Yao, X.~Huang, J.~Wang, J.~Yu, Q.~Guo, Y.~Yu,
  Y.~Zhang, J.~Wang, H.~Tao, D.~Yan, Z.~Yi, F.~Peng, F.~Jiang, H.~Zhang,
  L.~Deng, Y.~Zhang, Z.~Lin, C.~Zhang, S.~Zhang, M.~Guo, S.~Gu, G.~Fan,
  Y.~Wang, X.~Jin, Q.~Liu, and Y.~Tian, ``{PanGu-$\alpha$: Large-scale
  Autoregressive Pretrained Chinese Language Models with Auto-parallel
  Computation},'' 2021.

\bibitem{ref7}
D.~Lepikhin, H.~Lee, Y.~Xu, D.~Chen, O.~Firat, Y.~Huang, M.~Krikun, N.~Shazeer,
  and Z.~Chen, ``{GShard: Scaling Giant Models with Conditional Computation and
  Automatic Sharding},'' {\em arXiv}, 2020.

\bibitem{ref8}
W.~Fedus, B.~Zoph, and N.~Shazeer, ``{Switch Transformers: Scaling to Trillion
  Parameter Models with Simple and Efficient Sparsity},'' pp.~1--31, 2021.

\bibitem{ref9}
Z.~Zhang, X.~Han, H.~Zhou, P.~Ke, Y.~Gu, D.~Ye, Y.~Qin, Y.~Su, H.~Ji, J.~Guan,
  F.~Qi, X.~Wang, Y.~Zheng, G.~Zeng, H.~Cao, S.~Chen, D.~Li, Z.~Sun, Z.~Liu,
  M.~Huang, W.~Han, J.~Tang, J.~Li, X.~Zhu, and M.~Sun, ``{CPM: A Large-scale
  Generative Chinese Pre-trained Language Model},'' 2020.

\bibitem{ref10}
Z.~Zhang, Y.~Gu, X.~Han, S.~Chen, C.~Xiao, and Z.~Sun, ``{CPM-2 : Large-scale
  Cost-efficient Pre-trained Language Models},'' no.~2, 2021.

\bibitem{ref11}
J.~Lin, R.~Men, A.~Yang, C.~Zhou, M.~Ding, Y.~Zhang, P.~Wang, A.~Wang,
  L.~Jiang, X.~Jia, J.~Zhang, J.~Zhang, X.~Zou, Z.~Li, X.~Deng, J.~Liu, J.~Xue,
  H.~Zhou, J.~Ma, J.~Yu, Y.~Li, W.~Lin, J.~Zhou, J.~Tang, and H.~Yang, ``{M6: A
  Chinese Multimodal Pretrainer},'' mar 2021.

\bibitem{ref28}
N.~Alarcon, ``Openai presents gpt-3, a 175 billion parameters language model.''
  \url{https://developer.nvidia.com/blog/openai-presents-gpt-3-a-175-billion-parameters-language-model/}.

\bibitem{ref12}
A.~Radford, K.~Narasimhan, T.~Salimans, and I.~Sutskever, ``{Improving Language
  Understanding by Generative Pre-Training},'' {\em Encyclopedia of Autism
  Spectrum Disorders}, pp.~2640--2640, 2021.

\bibitem{ref13}
I.~S. {Alec Radford, Jeffrey Wu, Rewon Child, David Luan, Dario Amodei},
  ``{Language Models are Unsupervised Multitask Learners},'' {\em OpenAI Blog},
  vol.~1, no.~May, pp.~1--7, 2020.

\bibitem{ref14}
M.~Shoeybi, M.~Patwary, R.~Puri, P.~LeGresley, J.~Casper, and B.~Catanzaro,
  ``{Megatron-LM: Training Multi-Billion Parameter Language Models Using Model
  Parallelism},'' 2019.

\bibitem{ref15}
T.~Chen, B.~Xu, C.~Zhang, and C.~Guestrin, ``{Training Deep Nets with Sublinear
  Memory Cost},'' 2016.

\bibitem{ref16}
D.~Narayanan, M.~Shoeybi, J.~Casper, P.~LeGresley, M.~Patwary, V.~A.
  Korthikanti, D.~Vainbrand, P.~Kashinkunti, J.~Bernauer, B.~Catanzaro,
  A.~Phanishayee, and M.~Zaharia, {\em {Efficient Large-Scale Language Model
  Training on GPU Clusters}}, vol.~1.
\newblock Association for Computing Machinery, 2021.

\bibitem{ref17}
Y.~You, J.~Li, S.~Reddi, J.~Hseu, S.~Kumar, S.~Bhojanapalli, X.~Song,
  J.~Demmel, K.~Keutzer, and C.-J. Hsieh, ``{Large Batch Optimization for Deep
  Learning: Training BERT in 76 minutes},'' 2019.

\bibitem{ref18}
L.~Xu, H.~Hu, X.~Zhang, L.~Li, C.~Cao, Y.~Li, Y.~Xu, K.~Sun, D.~Yu, C.~Yu,
  Y.~Tian, Q.~Dong, W.~Liu, B.~Shi, Y.~Cui, J.~Li, J.~Zeng, R.~Wang, W.~Xie,
  Y.~Li, Y.~Patterson, Z.~Tian, Y.~Zhang, H.~Zhou, S.~Liu, Z.~Zhao, Q.~Zhao,
  C.~Yue, X.~Zhang, Z.~Yang, K.~Richardson, and Z.~Lan, ``{CLUE: A Chinese
  Language Understanding Evaluation Benchmark},'' apr 2020.

\bibitem{ref19}
Y.~Sun, S.~Wang, S.~Feng, S.~Ding, C.~Pang, J.~Shang, J.~Liu, X.~Chen, Y.~Zhao,
  Y.~Lu, W.~Liu, Z.~Wu, W.~Gong, J.~Liang, Z.~Shang, P.~Sun, W.~Liu, X.~Ouyang,
  D.~Yu, H.~Tian, H.~Wu, and H.~Wang, ``{ERNIE 3.0: Large-scale Knowledge
  Enhanced Pre-training for Language Understanding and Generation},'' 2021.

\bibitem{ref20}
C.~Zheng, M.~Huang, and A.~Sun, ``{ChID: A Large-scale Chinese IDiom Dataset
  for Cloze Test},'' jun 2019.

\bibitem{ref21}
L.~Xu, X.~Lu, C.~Yuan, X.~Zhang, H.~Xu, H.~Yuan, G.~Wei, X.~Pan, X.~Tian,
  L.~Qin, and H.~Hai, ``{FewCLUE: A Chinese Few-shot Learning Evaluation
  Benchmark},'' jul 2021.

\bibitem{ref22}
Y.~Chang, M.~Narang, H.~Suzuki, G.~Cao, J.~Gao, and Y.~Bisk, ``{WebQA: Multihop
  and Multimodal QA},'' sep 2021.

\bibitem{ref23}
Y.~Cui, T.~Liu, W.~Che, L.~Xiao, Z.~Chen, W.~Ma, S.~Wang, and G.~Hu, ``{A
  Span-Extraction Dataset for Chinese Machine Reading Comprehension},'' oct
  2018.

\bibitem{ref24}
T.~Z. Zhao, E.~Wallace, S.~Feng, D.~Klein, and S.~Singh, ``{Calibrate Before
  Use: Improving Few-Shot Performance of Language Models},'' feb 2021.

\bibitem{ref25}
A.~Holtzman, P.~West, V.~Shwartz, Y.~Choi, and L.~Zettlemoyer, ``{Surface Form
  Competition: Why the Highest Probability Answer Isn't Always Right},'' apr
  2021.

\bibitem{ref26}
Y.~Song, S.~Shi, J.~Li, and H.~Zhang, ``{Directional skip-gram: Explicitly
  distinguishing left and right context forword embeddings},'' {\em NAACL HLT
  2018 - 2018 Conference of the North American Chapter of the Association for
  Computational Linguistics: Human Language Technologies - Proceedings of the
  Conference}, vol.~2, pp.~175--180, 2018.

\bibitem{ref29}
Y.~Wang, P.~Ke, Y.~Zheng, K.~Huang, Y.~Jiang, X.~Zhu, and M.~Huang, ``{A
  Large-Scale Chinese Short-Text Conversation Dataset},'' aug 2020.

\end{thebibliography}
\end{document}